\documentclass[sigconf]{acmart}
\pdfoutput=1
\usepackage{color, colortbl}
\usepackage{natbib}
\definecolor{Gray}{gray}{0.9}
\definecolor{LightCyan}{rgb}{0.88,1,1}
\usepackage{microtype}

\AtBeginDocument{%
  \providecommand\BibTeX{{%
    \normalfont B\kern-0.5em{\scshape i\kern-0.25em b}\kern-0.8em\TeX}}}

\copyrightyear{2021}
\acmYear{2021}
\setcopyright{rightsretained}
\acmConference[ICMR '21]{Proceedings of the 2021 International Conference on Multimedia Retrieval}{August 21--24, 2021}{Taipei, Taiwan}
\acmBooktitle{Proceedings of the 2021 International Conference on Multimedia Retrieval (ICMR '21), August 21--24, 2021, Taipei, Taiwan}

\acmDOI{10.1145/3460426.3463585}
\acmISBN{978-1-4503-8463-6/21/08}

\acmSubmissionID{108}

\setcopyright{rightsretained}
\begin{document}
\fancyhead{}

\title{Evaluating Contrastive Models for Instance-based Image Retrieval}

\author{Tarun Krishna, Kevin McGuinness and Noel O'Connor}
\email{tarun.krishna2@mail.dcu.ie}
\affiliation{%
  \institution{SFI Insight Centre for Data Analytics, Dublin City University, Ireland}
  \country{}
}

\begin{abstract}
  In this work, we evaluate contrastive models for the task of image retrieval. We hypothesise that models that are learned to encode semantic similarity among  instances via discriminative learning should perform well on the task of image retrieval, where relevancy is defined in terms of instances of the same object. Through our extensive evaluation, we find that representations from models trained using contrastive methods perform on-par with (and outperforms) a pre-trained supervised baseline trained on the ImageNet labels in retrieval tasks under various configurations. This is remarkable given that the contrastive models require no explicit supervision. Thus, we conclude that these models can be used to bootstrap base models to build more robust image retrieval engines.   
\end{abstract}



\maketitle

\section{Introduction}
Large scale image retrieval, where the task is to search a large image collection for the most relevant image/content for a given query, is a fundamental task in computer vision. Since their inception, convolutional neural networks (ConvNets) \cite{krizhevsky2012imagenet, simonyan2014very} have become the prominent approach for extracting descriptors for image retrieval.  These descriptors perform very well in capturing the global semantics of an image and this has led to  state-of-the-art results on many benchmark computer vision tasks \cite{ren2016faster,chen2014semantic,he2016deep}.

The activations in the intermediate layers in ConvNets can be  used as a descriptor for an image. These descriptors are often  followed by some encoding methods for a compact representation. These encoding techniques range from  traditional approaches of VLAD \cite{jegou2010aggregating}, BoW \cite{mohedano2016bags}, and Fisher vectors \cite{perronnin2007fisher}, to simple pooling methods like Maximum Activation of Convolution (MAC) \cite{Azizpour_2015_CVPR_Workshops}, Sum Pooling of Convolution (SPoC) \cite{babenko2015aggregating}, Regional-MAC \cite{tolias2015particular}, etc. The drawback of these methods is that these (off-the-shelf) network are trained  to reduce inter-class variance through supervision on ImageNet classes, this might affect the performance of instance retrieval (i.e.~retrieving images that represent the same object or scene as in a query), which is a more fine-grained task.

This drawback has been addressed in the literature by fine-tuning \cite{gordo2016deep, radenovic2016cnn, gordo2017end, DBLP:journals/corr/abs-1906-07589}. We hypothesise that a simpler approach could be to retain a traditional off-the-shelf regime but instead use  models that are trained based on instance-wise supervision using similarity-based learning. To this end, we investigate contrastive learning based methods, i.e.~trained in an unsupervised fashion using contrastive loss \cite{wu2018unsupervised,gutmann2010noise}.
This learning regime relies on learning a meaningful embedding that captures inherent similarity between  instances using discriminative approaches \cite{wu2018unsupervised}. This work investigates the effectiveness of contrastive methods that capture this very idea of instance similarity. To summarize our contributions:
\vspace{-0.4em}
\begin{itemize}
    \item we extensively evaluate contrastive methods as a fixed feature extractor across different benchmark;
    \item we provide experimental evidence showing that these models (trained without any explicit supervision) perform on par with a pre-trained supervised baseline (Table \ref{tab1} and \ref{tab2}); 
    \item we further investigate the role of the dimensionality of the feature embeddings for this task (Table \ref{tab3}). 
\end{itemize}


\section{Related Work}

Conventional image retrieval methods \cite{sivic2003video, nister2006scalable}  relied on bag-of-words models that exploit local invariant features such as SIFT  \cite{lowe2004distinctive} and large visual vocabularies (e.g.~\cite{philbin2007object}). To aggregate local patches
and build a global summary, encoding methods such as Fisher vectors \cite{perronnin2007fisher} or VLAD \cite{jegou2010aggregating}, have also been proposed \cite{perronnin2010large,gordoa2012leveraging,radenovic2015multiple}.

Since the introduction of Deep-ConvNets, \cite{krizhevsky2012imagenet,simonyan2014very,donahue2014decaf,he2016deep} there has been a paradigm shift to exploit deep features instead of hand-crafted ones. Intermediate layers in convolutional nets can be used as global or local descriptors.  As a result, so-called off-the-shelf \cite{azizpour2015generic, razavian2014cnn} features can be used for retrieval.  Based on this authors in \cite{babenko2015aggregating} used sum pooling with a centre prior for aggregating features across spatial dimensions. Other conventional encoding techniques like VLAD \cite{gong2014multi} or Fisher kernels \cite{perronnin2015fisher} have also been used in combination with these local feature maps. For example, in \cite{mohedano2016bags} proposed BoW-encodings of  convolutional features for instance retrieval, whilst
\cite{tolias2015particular}  proposed R-MAC using max activations over a grid of windows of different scales to obtain compact representations. 

Most of the off-the-shelf features are trained on ImageNet \cite{russakovsky2015imagenet} to reduce inter-class variance. However, this may degrade the performance of an instance-based retrieval system. One way to address this is to finetune the model as shown in \cite{babenko2014neural, gordo2016deep, radenovic2016cnn, gordo2017end,DBLP:journals/corr/abs-1906-07589}. In the context of image retrieval, most of the finetuning has been performed on \textit{Landmark} datasets \cite{babenko2014neural}, which further requires cleaning of non-related images and potentially expensive post-processing. 

Another way is to exploit methods that are trained to reduce intra-class variance, as is the case in  contrastive learning.  Unlike in supervised learning, these approaches learn to discriminate among individual instances without any concept of categories. This work \cite{wu2018unsupervised} discusses this notion of instance discrimination. Building on this, a simple formulation is presented in SimCLR \cite{pmlr-v119-chen20j}. The intuition behind these approaches is to maximize the agreement among augmented views of the same instance using Noise Contrastive Estimation (NCE) \cite{gutmann2010noise}.  Minimising NCE is equivalent to maximizing mutual information (MI) as was formally shown in CPC \cite{oord2018representation} as InfoNCE. DIM \cite{hjelm2018learning} and AMDIM \cite{bachman2019learning} further extend the idea of InfoNCE across multiple views and scales. One of the downsides of these approaches is that they require large batch sizes on large GPU clusters. To address this drawback, the authors in \cite{he2020momentum, chen2020improved} introduced MoCo, which uses an online and  momentum updated offline network that views contrastive learning as a dictionary lookup task. Intuitively, the ability to  discriminate among individual instances inherently encoded through the learning makes contrastive learning a good candidate for the task of instance retrieval. 
\section{Contrastive Models}

Contrastive learning refers to learning by comparison. This comparison is performed between positive pairs of ``similar'' and negative pairs of ``dissimilar'' inputs, which is achieved via a contrastive loss \cite{pmlr-v119-chen20j, le2020contrastive} derived from  Noise Contrastive Estimation (NCE) \cite{gutmann2010noise}.
%
%

 This work targets representative contrastive models for investigation, which are diverse in terms of the way they fuse information. Intuitively, this should lead to each model spanning different feature spaces from an image understanding perspective. The following briefly describes the models considered in this study. 

    \noindent \textbf{AMDIM}: \textit{Augmented Multiscale DIM} \cite{bachman2019learning} extends the Deep Info-Max (DIM) framework \cite{hjelm2018learning}  by learning features extracted from multiple views of a shared context.
     Local DIM maximizes MI across features extracted independently from augmented views of each input along with features across multiple scales with more powerful encoders. 
     
     \noindent \textbf{MoCo}: \textit{Momentum Contrast Representation Learning} \cite{he2020momentum, chen2020improved} alleviates the need for storing offline representations of the entire dataset in  memory \cite{wu2018unsupervised} through the use of a \textit{dynamic} memory \textit{queue}.  The samples in the dictionary are progressively replaced. 
 This approach looks at contrastive learning as a way of building a discrete dictionary (\textit{queue}) of inputs (data samples) for a high-dimensional space. We consider MoCo$_{v1}$ \cite{he2020momentum} and MoCo$_{v2}$ \cite{chen2020improved}.
    
    \noindent \textbf{SimCLR}: \textit{Simple framework for Contrastive Learning} \cite{pmlr-v119-chen20j} is a simplified framework for contrastive learning compared to the previous ones. Here stochastic data augmentations are applied on an input $\mathbf{x}$ to get two views of $\mathbf{x}_{i}$ and $\mathbf{x}_{j}$. Sequentially these augmented inputs are passed through a \textit{base encoder} $\mathbf{f(.)}$ followed by \textit{projection head}, a small network that projects representations from $\mathbf{f(.)}$ to space where a contrastive loss is applied. We investigate SimClr$_{1x, 2x, 4x}$.

\begin{table*}[t]
  \caption{Comparing mAP (\%) score across different models. Bold (red) best performing ensemble if it exists.}
  \label{tab1}
  \begin{tabular}{ll|c|c|c|c|c|c|c|c}
    \hline
      && \textbf{G}& \textbf{G} & \textbf{G} & \textbf{G} &\textbf{G}&\textbf{G}&\textbf{G}&\textbf{G}\\
      &&&+AQE & +DFS  & +DBA & +AQE & +DBA&+DBA&+DBA\\
    &&&&&&+DFS&+AQE&+DFS&+AQE \\
    Dataset&Method&&&&&&&&+DFS \\
    \hline
    \hline
    &Baseline & 55.12 & 67.85& 74.29 & 62.63 & \textbf{82.64} &70.6 & 77.95 & \textbf{80.4}\\
    \rowcolor{Gray}
    &SimClr$_{1x}$& 51.47 & 62.96 & 71.42 & 61.00 & 76.40 & 66.32 & 72.98 & 73.38\\
    &SimClr$_{2x}$& 58.59 & \textbf{70.49} & 79.11 & 67.34 & 82.34 & \textbf{72.62} & \textbf{79.02} & 79.96\\
    \rowcolor{Gray}
    &SimClr$_{4x}$& \textbf{59.40} & 69.80& \textbf{80.07} & \textbf{67.52} & 82.00 &71.03 & 78.67 & 76.98 \\
  &MoCo$_{v1}$ &56.76& 66.89 & 74.78& 64.12 & 76.44& 69.77 & 73.39 & 73.95\\
   \rowcolor{Gray}
   \textbf{Oxford 5k} &MoCo$_{v2}$ &58.36 & 67.49 & 75.26 & 65.09 & 78.07& 69.86 & 72.52 & 72.86\\
    &AmDim & 36.95 & 43.44 & 48.13 & 39.97 & 52.99 & 44.69 & 46.3 & 47.74\\
    \rowcolor{Gray}
    &SimClr$_{2x}$+SimClr$_{4x}$&\textcolor{red}{\textbf{61.89}}&\textcolor{red}{\textbf{72.64}} &\textcolor{red}{\textbf{81.66}} & \textcolor{red}{\textbf{69.27}} & \textcolor{red}{\textbf{86.02}}& \textcolor{red}{\textbf{74.43}} & \textcolor{red}{\textbf{80.45}} & \textcolor{red}{\textbf{81.99}} \\
    &SimClr$_{2x}$+AmDim& 57.06 & 67.37 & 75.63 & 64.69&77.77 & 69.29&72.84&74.51 \\
    \rowcolor{Gray}
    &SimClr$_{2x}$+MoCo$_{v2}$& 61.33 & 71.58&78.97&68.47&82.33&72.94&78.36&79.03 \\
    &SimClr$_{4x}$+MoCo$_{v2}$ &61.87 & 71.97 &80.86&68.20&84.02&72.94&78.56&79.57\\
    \rowcolor{Gray}
    &SimClr$_{4x}$+AmDim &58.13 &68.76 &77.66 &64.81 &80.05 &69.78&75.93&77.32 \\
    \rowcolor{LightCyan}
    &GeM (AP)& 66.90 & 78.57 & 90.36 & 75.70& 94.34& 81.92 & 89.65 & 91.12\\
    \hline
    
    &Baseline & 41.46 & 61.76 & 83.74 & 64.71 & 90.17 & \textbf{75.56} &90.17 & 90.62\\
    \rowcolor{Gray}
    &SimClr$_{1x}$ & 42.28 & 61.15& 85.16 & 64.36 & 90.17& 74.45 & 89.37 & 90.49\\
    
    &SimClr$_{2x}$ & 44.48 & 59.65 & \textbf{89.39} & 63.28 & \textbf{91.61} & 71.91& \textbf{90.82} & 
    
    \textbf{91.48}\\
    \rowcolor{Gray}
    &SimClr$_{4x}$ & 44.96 & 56.3 & 88.98 & 60.40 & 90.18 & 68.46 & 88.18 & 88.74\\
  &MoCo$_{v1}$& 43.22 & 60.34 & 83.13 & 63.00& 86.9 & 72.11 & 87.91 & 88.73\\
  \rowcolor{Gray}
    \textbf{Paris 6k}&MoCo$_{v2}$& \textbf{49.72} & \textbf{64.46} & 88.02 & \textbf{67.06} & 89.86 & 75.01 & 88.65 & 89.01\\
    &AmDim & 34.87 & 49.2 & 59.55 & 47.72 & 63.7 & 56.06 & 64.18 & 65.08\\
   \rowcolor{Gray}
    &SimClr$_{2x}$+SimClr$_{4x}$&45.27&56.84 & 89.97&59.88&91.60&68.07&89.19&90.28 \\
    &SimClr$_{2x}$+AmDim & 44.45 & 60.06 & 86.71 &63.02 & 89.06&71.12 & 88.46 & 88.76\\
    \rowcolor{Gray}
    &SimClr$_{2x}$+MoCo$_{v2}$ &47.93 & 61.46&\textcolor{red}{\textbf{90.92}} & 64.96&\textcolor{red}{\textbf{92.47}}&72.37&90.42&91.08\\
    &SimClr$_{4x}$+MoCo$_{v2}$&47.44&58.44&90.40&61.88&91.97&69.48&89.00&89.93 \\
    \rowcolor{Gray}
    &SimClr$_{4x}$+AmDim &44.74 & 57.05 & 87.59&60.66&88.97&68.60&87.06 &87.45\\
    \rowcolor{LightCyan}
    &GeM (AP)& 49.87&63.46 & 93.99 & 67.99 & 95.56 & 76.78 & 90.15 & 94.07\\
    \hline
    
    &Baseline& 36.64 & \textbf{64.42} & 71.7 & \textbf{63.96} & \textbf{76.43} & \textbf{74.32} & 16.41 & \textbf{76.13}\\
    \rowcolor{Gray}
    &SimClr$_{1x}$ & 27.51 & 47.85 & 55.38 & 47.34 & 60.61 & 57.73 & 21.00 & 61.15\\
    &SimClr$_{2x}$ & 36.68 & 57.04 & 66.47 & 56.89 & 69.48 & 65.27 & 24.93 & 67.18 \\
    \rowcolor{Gray}
    &SimClr$_{4x}$ & \textbf{44.76} & 63.73 & \textbf{73.51} & 63.26 & 75.02 & 70.17 & 28.79 & 70.54\\
  &MoCo$_{v1}$ & 33.44 & 54.88 & 66.31 & 54.84 & 70.1 & 64.65 & 29.25 & 69.8\\
    \rowcolor{Gray}
    \textbf{INSTRE}&MoCo$_{v2}$ & 33.36 & 51.79 & 60.66 & 50.47 & 62.56 & 58.83 & 22.38 & 59.77\\
    &AmDim& 24.34 & 37.92& 45.47 & 36.69 & 47.14 & 42.74 & \textbf{29.47}& 47.71\\
    \rowcolor{Gray}
    &SimClr$_{2x}$+SimClr$_{4x}$&45.58&63.94&73.54&63.54&74.81 &70.54&28.73&70.75 \\
    &SimClr$_{2x}$+AmDim & 39.37& 59.02&68.85&59.09&71.14 &66.68&26.19&67.93\\
    \rowcolor{Gray}
    &SimClr$_{2x}$+MoCo$_{v2}$ & 41.99&60.99&70.42&60.64&72.24&68.41&26.09&68.69\\
    &SimClr$_{4x}$+MoCo$_{v2}$&\textcolor{red}{\textbf{47.07}}&\textcolor{red}{\textbf{65.23}} & \textcolor{red}{\textbf{74.64}}& \textcolor{red}{\textbf{64.93}}&76.12&71.42 & 27.09&71.26 \\
    \rowcolor{Gray}
    &SimClr$_{4x}$+AmDim& 45.83 &63.88 & 74.25 &63.84&75.28&70.83&\textcolor{red}{\textbf{30.07}}&71.60 \\
    \rowcolor{LightCyan}
    &GeM (AP)& 20.78 &31.37 & 34.85 & 30.29 & 38.06 & 35.93 &18.48 & 37.54\\
  \bottomrule
\end{tabular}
\end{table*}
\begin{table*}[t]
  \caption{Comparison on global ranking across different model. Bold (red) best performing ensemble if it exists.}
  \label{tab2}
  \begin{tabular}{ll| ccc| ccc |cccl}
    \hline
    \multicolumn{1}{l}{}&\multicolumn{1}{l|}{}&\multicolumn{3}{c|}{\textbf{Easy}}&\multicolumn{3}{c|}{\textbf{Medium}}& \multicolumn{3}{c}{\textbf{Hard}}\\
  
    \multicolumn{1}{l}{Dataset}&\multicolumn{1}{l|}{Method}&  mAP & mp@5 &mp@10 & mAP  & mp@5&mp@10 & mAP  &mp@5 &mp@10\\
    \hline
    \hline
    &Baseline & 45.65 & 67.55 &61.81 &32.89 &64.00 &58.57 & 12.10 &23.71&19.43\\
     \rowcolor{Gray}
    &SimClr$_{1x}$ &47.17 &69.78&63.90&31.98&65.24&57.52&9.52&20.57&15.86 \\
    &SimClr$_{2x}$ &\textbf{54.95}&\textbf{76.69}&\textbf{69.49}&38.54&73.14&\textbf{65.33}&\textbf{14.41}&\textbf{31.21}&\textbf{22.93} \\
     \rowcolor{Gray}
    &SimClr$_{4x}$ &54.65&74.80&68.37&\textbf{38.57}&\textbf{73.71}&65.05&14.05&30.07&22.79\\
    &MoCo$_{v1}$ &48.29&68.48&64.36&33.57&64.43&58.58&9.27&20.05&15.62 \\
     \rowcolor{Gray}
    \textbf{rOxford 5k} &MoCo$_{v2}$&52.69&72.33&65.86&36.49&67.24&59.24&10.72&23.38&18.67 \\
    &AmDim & 21.24 &38.31 &31.35&17.54&36.38&32.57&4.11 &5.92&6.33\\
    \rowcolor{Gray}
    &SimClr$_{2x}$+SimClr$_{4x}$ & \textcolor{red}{\textbf{55.23}} & 76.15 & 68.87&39.60&\textcolor{red}{\textbf{74.29}} & \textcolor{red}{\textbf{65.64}}&15.01 & 29.43&\textcolor{red}{\textbf{23.97}}\\
    &SimClr$_{2x}$+AmDim& 52.08&72.28&65.51&36.19&66.86&58.57&10.20&21.64&17.64 \\
    \rowcolor{Gray}
    &SimClr$_{2x}$+MoCo$_{v2}$&55.11& 76.47&69.35&38.46&70.52&62.57&12.51&27.12&21.83 \\
    &SimClr$_{4x}$+MoCo$_{v2}$ & 53.98 &72.40&67.90&38.31&70.10&64.69&13.38&28.43&23.13\\
    \rowcolor{Gray}
    &SimClr$_{4x}$+AmDim& 49.23&70.76&64.00&35.72&67.62&61.33&11.72&23.79&18.80 \\
    \rowcolor{LightCyan}
    &GeM (AP)& 64.07 & 84.93 & 80.56 & 51.03&89.43&83.86&30.30 & 54.86 & 44.00 \\

    \hline
    &Baseline & 47.20&91.14&87.00&31.63&93.71&91.71&10.44&57.14&43.43\\
     \rowcolor{Gray}
    &SimClr$_{1x}$ & 49.35&93.14&89.52&31.61&95.71&92.00&8.93&46.57&38.43\\
    &SimClr$_{2x}$ &54.28&\textbf{94.00}&91.14&34.47&96.00&92.86&10.87&62.00&49.71 \\
     \rowcolor{Gray}
    &SimClr$_{4x}$ &54.97&93.71&\textbf{91.71}&35.39&\textbf{96.29}&\textbf{94.86}&12.34&\textbf{66.57}&\textbf{56.29}\\
  &MoCo$_{v1}$&50.47&92.29&89.71&31.48&95.14&92.57&8.13&46.57&36.29\\
     \rowcolor{Gray}
   \textbf{rParis 6k} &MoCo$_{v2}$&\textbf{55.77}&92.86&89.24&\textbf{36.32}&94.86&92.14&\textbf{11.71}&60.00&50.43\\
    &AmDim & 38.62&80.29&73.71&25.92&82.57&75.71&6.36&25.14&19.57\\
    \rowcolor{Gray}
    &SimClr$_{2x}$+SimClr$_{4x}$& 55.54 & 94.00&\textcolor{red}{\textbf{91.86}}&35.59&96.00&94.00&12.37&66.29&56.00 \\
    &SimClr$_{2x}$+AmDim& 53.71 & 93.71 & 90.43&34.57&95.14&93.14&10.46&55.71 & 46.00
\\
\rowcolor{Gray}
&SimClr$_{2x}$+MoCo$_{v2}$ & \textcolor{red}{\textbf{57.00}} & 93.43&91.00&\textcolor{red}{\textbf{36.66}}&95.14 &93.29&11.88&64.00&52.43\\
    &SimClr$_{4x}$+MoCo$_{v2}$ & 56.86 & \textcolor{red}{\textbf{94.29}} & 91.71 & 36.61 & 96.29 & 94.86& \textcolor{red}{\textbf{12.68}}& \textcolor{red}{\textbf{69.43}}&55.57\\
    \rowcolor{Gray}
    &SimClr$_{4x}$+AmDim& 54.65&92.86 & 90.57 & 35.06 & 95.71 & 93.43&11.41 &61.14& 48.71 \\
    \rowcolor{LightCyan}
    &GeM (AP)&54.90 & 94.38 & 91.95&37.36 & 98.86&97.00& 14.65 &76.86 & 63.71\\
  \bottomrule
\end{tabular}
\end{table*}

\begin{table}[t]
  \caption{Comparison of mAP (\%) across different PCA dimension and the true dimension.}
  \label{tab3}
  \resizebox{\columnwidth}{!}{%
  \begin{tabular}{ll| ccccccc|ccc}
    \hline
    \multicolumn{1}{l}{}&\multicolumn{1}{l|}{}&\multicolumn{7}{c|}{\textbf{PCA-Whitening}}& \multicolumn{1}{c}{\textbf{True}}\\
    \multicolumn{1}{l}{Dataset}&\multicolumn{1}{l|}{Method}&  32 & 64 & 128 & 256 & 512 & 1024 & 2048 & \textbf{dim.} \\
    \hline
    \hline
    &Baseline & 48.31& 54.61 & 56.68 & 57.52 & 55.12& 49.44 &37.23 &\textbf{58.47}\\
    \rowcolor{Gray}
    &SimClr$_{1x}$ & 34.96 & 44.96 & 54.17 & \textbf{54.79}& 51.47&44.16 &34.94 &50.63 \\
    &SimClr$_{2x}$ & 36.79 & 47.96 & 60.32 & \textbf{61.68}&58.59&49.82& 24.62& 53.72\\
    \rowcolor{Gray}
    \textbf{Oxford 5k}&SimClr$_{4x}$ &35.32 & 45.86 & 57.77 & \textbf{61.70}&59.40& 53.83& 44.71&41.94\\
    &MoCo$_{v1}$ & 29.90 & 41.95 & 54.62 & \textbf{57.38}&56.76&50.20& 39.37&38.73\\
    \rowcolor{Gray}
    &MoCo$_{v2}$& 39.92 & 47.88 & 57.92&\textbf{60.43}&58.36&52.14& 40.05&51.76\\
    &AmDim & 11.52 & 15.30 & 21.71 & 28.71&36.95& \textbf{37.58}& 30.73&16.10\\

    \hline
    &Baseline &\textbf{72.22} & 71.83& 63.65 & 53.40& 41.46&28.99 & 17.89&68.36\\
    \rowcolor{Gray}
    &SimClr$_{1x}$ & 69.90 & \textbf{73.29} & 66.62& 54.58&42.28&30.7& 20.06&66.60\\
    &SimClr$_{2x}$ & 75.21 & \textbf{77.90}& 69.25&57.83&44.48&33.05&23.27&72.20\\
    \rowcolor{Gray}
    \textbf{Paris 6k}&SimClr$_{4x}$ &77.16&\textbf{78.04} & 69.19&57.29&44.96&35.02&25.79&72.89\\
    &MoCo$_{v1}$&56.84&\textbf{63.63}&61.87&54.07 &43.22&32.57&21.63&53.66\\
    \rowcolor{Gray}
    &MoCo$_{v2}$&70.08& \textbf{75.11}& 71.24& 61.66&49.72&35.73&22.98 &69.99\\
    &AmDim & 21.49 & 33.96 & \textbf{41.19} & 40.85&34.87& 26.82& 17.85&25.50\\
    \hline
    &Baseline &26.44& 34.92 & \textbf{38.68}& 38.55&36.64&29.25 &20.20 &33.03\\
    \rowcolor{Gray}
    &SimClr$_{1x}$& 16.47 & 23.27 & 29.01&\textbf{30.35}&27.51&21.85& 15.97&21.85\\
    &SimClr$_{2x}$&18.42 & 26.83 & 35.88& \textbf{39.67}&36.68 &29.64&21.57&25.94\\
    \rowcolor{Gray}
    \textbf{INSTRE}&SimClr$_{4x}$& 18.92& 28.97 & 40.45& \textbf{46.99}&44.76&36.68&27.65&28.98\\
    &MoCo$_{v1}$& 20.08 & 27.85 & 33.77& \textbf{36.24}&33.44&26.66& 18.46&23.01\\
\rowcolor{Gray}
    &MoCo$_{v2}$& 19.00 & 27.38& 34.34 & \textbf{36.22}&33.36&26.86& 18.79&26.22\\
    &AmDim&10.65 & 15.88& 20.55 & 24.15&\textbf{24.34}&20.45& 14.45&10.25\\
  \bottomrule
\end{tabular}}
\end{table}
\section{Experiments}
This section describes our experimental setup for the evaluation. 
\subsection{Setup}
We evaluate models on three standard benchmark datasets: Oxford5k \cite{philbin2007object}, Paris6k \cite{philbin2008lost}, and INSTRE \cite{wang2015instre}.
Retrieval performance is measured using mean Average Precision (mAP) following standard procedures for Oxford 5k and Paris 6k benchmarks and for INSTRE evaluating mAP over 1200 images as described in \cite{iscen2017efficient}.  We further evaluate the performance on revised rOxford 5k and rParis 6k using the new evaluation protocol based on \textit{easy}, \textit{medium}, and \textit{hard} ground truth labels  \cite{radenovic2018revisiting}. For the revised benchmarks we report both mAP and mean precision@(10,5) (mp@10, mp@5).

The goal is not to fine-tune the models but instead evaluate them as a fixed feature extractor to obtain visual descriptors. The base encoder of each of the models is some flavour of ResNet\footnote{apart from simClr$_{2x,4x}$ each uses ResNet50 as a backbone encoder} with varying complexity. To this end, we consider the output of the last convolutional layer, i.e.~just before the adaptive pooling layer, as our descriptor, which leads to feature maps of size $\mathbb{R}^{ C\times H \times W}$. To obtain compact representations  we use R-MAC ($L=3a $) \cite{tolias2015particular} over spatial dimensions to get a fixed representation of size $\mathbb{R}^C$. We further post-process the vectors by applying $L_{2}$ normalization, PCA-whitening, and $L_{2}$ normalization again. 

We resize our input images to a fixed resolution of $724\times 724$ giving a feature spatial dimension of $23 \times 23$ except in the case of AMDIM where the dimensions are $40 \times 40$. However, we downsample this to $23\times23$ to keep uniformity across the evaluation. 
Also, before running the final evaluation we first run each of the models in training mode (PyTorch \texttt{model.train()} just feed-forward) to tune the batch-normalization statistics to the current dataset and then finally test models in evaluation mode (\texttt{model.eval()}). 


\noindent\textbf{Baseline.} For comparing across all the contrastive models we use \textit{ResNet50} \cite{he2016deep} trained on \textit{ImageNet} as a fixed feature extractor as our pre-trained supervised baseline model.
\textbf{Note.} For completeness we also evaluate a fine-tuned model  \cite{DBLP:journals/corr/abs-1906-07589}, which uses Generalized Mean Pooling  (instead of R-MAC) trained with Average Precision loss (GeM (AP)) \footnote{\url{https://github.com/naver/deep-image-retrieval}}. The purpose of this is to provide an indicative upper bound to the evaluation scores.

\noindent \textbf{Ranking.} We consider global search (G) in this evaluation.  We further integrate Global search with Average Query Expansion \cite{chum2007total} (AQE), DataBase Augmentation (DBA), \cite{arandjelovic2012three} and Diffusion \cite{yang2019efficient} (DFS). For AQE we consider nearest neighbour $N=10$, for DBA we consider $N^{'}=20$ while combining both of these we consider $N=1$ and $N^{'}=20$, based on the findings in \cite{gordo2017end}.

We use a PCA dimension of 512 and  evaluate on a global search for R-MAC representations unless otherwise stated. 
\subsection{Results}
Table \ref{tab1} compares different models along with different expansion techniques\footnote{here and in Table~\ref{tab2} GeM (AP) serves as a upper bound indicator rather than a benchmark.}. For a naive global search on Oxford 5k, the contrastive approach achieves an mAP (\%) of 59.40 while the baseline achieves 55.12. Overall best performance is achieved with AQE and DFS for the baseline model (82.64) but this mAP score for SimCLR$_{2x}$ (82.34) is in the same range as the former. We also include results of ensembling  contrastive methods\footnote{The R-MAC representations are concatenated and dimensionally reduced via PCA}, which seems to give a further performance boost (in red in Table \ref{tab1}). A similar inference could be drawn for Paris 6k  where the best mAP for global is achieved by MoCO$_{v2}$ (49.72) while the overall best (91.61) is achieved with AQE and DFS for SimClr$_{2x}$. A further boost can be observed for the ensemble. In the case of the INSTRE dataset, we see a similar pattern for global
search with mAP 44.76 corresponding to SimClr$_{4x}$ while the overall best is achieved with AQE and DFS for the baseline (76.43) while the best result achieved for contrastive models is 75.02 for SimClr$_{4x}$. This clearly indicates that contrastive methods trained to reduce intra class variance capture the notion of instance similarity which is being reflected in this evaluation. Also expansion techniques further boosts the performance over global search.

To further consolidate our findings,  we also conducted an evaluation on the revised rOxford 5k and  rParis 6k datasets as depicted in Table \ref{tab2}. On  rOxford 5k SimClr$_{2x}$ gives the best performance on all labels.  mP@10 is almost 70\% for the \textit{easy} category, with the drop in performance for \textit{hard} label. Similarly, on rParis 6k, SimClr$_{4x}$ gives the best performance  with  mP@10 over 90 for easy and medium, but again this drops off for \textit{hard} labels. As in the Table \ref{tab1}, here ensembling further boosts performance. Again, contrastive models surpass the baseline pre-trained supervised model.

\noindent\textbf{Effect of descriptor dimension on performance}. Table \ref{tab3} reports our findings for global search\footnote{Comparison is across the horizontal dimension (columns)}. Interestingly, true dimensions ($L_2$ normalized R-MAC representations) appear to perform worse for almost all the models. The best dimension varies across the dataset but it is never the true dimension. 
This could be attributed to dimensions with small principal components being noisy and redundant and adversely affecting performance. 

\section{Conclusion}
This work evaluated contrastive models for the task of instance-based image retrieval. Our evaluation found that these methods are on par with those trained on class labels. In fact, in many settings in Table \ref{tab1}, \ref{tab2} contrastive approaches surpass the supervised model. The quantitative evaluation shows that these contrastive methods can easily surpass supervised models without any explicit supervision. 

\section{Acknowledgment}
This work has emanated from research supported by Science Foundation Ireland (SFI) under Grant Number SFI/12/RC/2289\_P2, co-funded by the European Regional Development Fund and Xperi FotoNation.

    
    

\bibliographystyle{ACM-Reference-Format}
\bibliography{sample-base}


\begin{thebibliography}{45}


\ifx \showCODEN    \undefined \def \showCODEN     #1{\unskip}     \fi
\ifx \showDOI      \undefined \def \showDOI       #1{#1}\fi
\ifx \showISBNx    \undefined \def \showISBNx     #1{\unskip}     \fi
\ifx \showISBNxiii \undefined \def \showISBNxiii  #1{\unskip}     \fi
\ifx \showISSN     \undefined \def \showISSN      #1{\unskip}     \fi
\ifx \showLCCN     \undefined \def \showLCCN      #1{\unskip}     \fi
\ifx \shownote     \undefined \def \shownote      #1{#1}          \fi
\ifx \showarticletitle \undefined \def \showarticletitle #1{#1}   \fi
\ifx \showURL      \undefined \def \showURL       {\relax}        \fi
\providecommand\bibfield[2]{#2}
\providecommand\bibinfo[2]{#2}
\providecommand\natexlab[1]{#1}
\providecommand\showeprint[2][]{arXiv:#2}

\bibitem[\protect\citeauthoryear{Arandjelovi{\'c} and
  Zisserman}{Arandjelovi{\'c} and Zisserman}{2012}]%
        {arandjelovic2012three}
\bibfield{author}{\bibinfo{person}{Relja Arandjelovi{\'c}} {and}
  \bibinfo{person}{Andrew Zisserman}.} \bibinfo{year}{2012}\natexlab{}.
\newblock \showarticletitle{Three things everyone should know to improve object
  retrieval}. In \bibinfo{booktitle}{\emph{2012 IEEE Conference on Computer
  Vision and Pattern Recognition}}. IEEE, \bibinfo{pages}{2911--2918}.
\newblock


\bibitem[\protect\citeauthoryear{Azizpour, Sharif~Razavian, Sullivan, Maki, and
  Carlsson}{Azizpour et~al\mbox{.}}{2015a}]%
        {Azizpour_2015_CVPR_Workshops}
\bibfield{author}{\bibinfo{person}{Hossein Azizpour}, \bibinfo{person}{Ali
  Sharif~Razavian}, \bibinfo{person}{Josephine Sullivan},
  \bibinfo{person}{Atsuto Maki}, {and} \bibinfo{person}{Stefan Carlsson}.}
  \bibinfo{year}{2015}\natexlab{a}.
\newblock \showarticletitle{From Generic to Specific Deep Representations for
  Visual Recognition}. In \bibinfo{booktitle}{\emph{Proceedings of the IEEE
  Conference on Computer Vision and Pattern Recognition (CVPR) Workshops}}.
\newblock


\bibitem[\protect\citeauthoryear{Azizpour, Sharif~Razavian, Sullivan, Maki, and
  Carlsson}{Azizpour et~al\mbox{.}}{2015b}]%
        {azizpour2015generic}
\bibfield{author}{\bibinfo{person}{Hossein Azizpour}, \bibinfo{person}{Ali
  Sharif~Razavian}, \bibinfo{person}{Josephine Sullivan},
  \bibinfo{person}{Atsuto Maki}, {and} \bibinfo{person}{Stefan Carlsson}.}
  \bibinfo{year}{2015}\natexlab{b}.
\newblock \showarticletitle{From generic to specific deep representations for
  visual recognition}. In \bibinfo{booktitle}{\emph{Proceedings of the IEEE
  conference on computer vision and pattern recognition workshops}}.
  \bibinfo{pages}{36--45}.
\newblock


\bibitem[\protect\citeauthoryear{Babenko and Lempitsky}{Babenko and
  Lempitsky}{2015}]%
        {babenko2015aggregating}
\bibfield{author}{\bibinfo{person}{Artem Babenko} {and} \bibinfo{person}{Victor
  Lempitsky}.} \bibinfo{year}{2015}\natexlab{}.
\newblock \showarticletitle{Aggregating deep convolutional features for image
  retrieval}.
\newblock \bibinfo{journal}{\emph{arXiv preprint arXiv:1510.07493}}
  (\bibinfo{year}{2015}).
\newblock


\bibitem[\protect\citeauthoryear{Babenko, Slesarev, Chigorin, and
  Lempitsky}{Babenko et~al\mbox{.}}{2014}]%
        {babenko2014neural}
\bibfield{author}{\bibinfo{person}{Artem Babenko}, \bibinfo{person}{Anton
  Slesarev}, \bibinfo{person}{Alexandr Chigorin}, {and} \bibinfo{person}{Victor
  Lempitsky}.} \bibinfo{year}{2014}\natexlab{}.
\newblock \showarticletitle{Neural codes for image retrieval}. In
  \bibinfo{booktitle}{\emph{European conference on computer vision}}. Springer,
  \bibinfo{pages}{584--599}.
\newblock


\bibitem[\protect\citeauthoryear{Bachman, Hjelm, and Buchwalter}{Bachman
  et~al\mbox{.}}{2019}]%
        {bachman2019learning}
\bibfield{author}{\bibinfo{person}{Philip Bachman}, \bibinfo{person}{R~Devon
  Hjelm}, {and} \bibinfo{person}{William Buchwalter}.}
  \bibinfo{year}{2019}\natexlab{}.
\newblock \showarticletitle{Learning representations by maximizing mutual
  information across views}.
\newblock \bibinfo{journal}{\emph{arXiv preprint arXiv:1906.00910}}
  (\bibinfo{year}{2019}).
\newblock


\bibitem[\protect\citeauthoryear{Chen, Papandreou, Kokkinos, Murphy, and
  Yuille}{Chen et~al\mbox{.}}{2014}]%
        {chen2014semantic}
\bibfield{author}{\bibinfo{person}{Liang-Chieh Chen}, \bibinfo{person}{George
  Papandreou}, \bibinfo{person}{Iasonas Kokkinos}, \bibinfo{person}{Kevin
  Murphy}, {and} \bibinfo{person}{Alan~L Yuille}.}
  \bibinfo{year}{2014}\natexlab{}.
\newblock \showarticletitle{Semantic image segmentation with deep convolutional
  nets and fully connected crfs}.
\newblock \bibinfo{journal}{\emph{arXiv preprint arXiv:1412.7062}}
  (\bibinfo{year}{2014}).
\newblock


\bibitem[\protect\citeauthoryear{Chen, Kornblith, Norouzi, and Hinton}{Chen
  et~al\mbox{.}}{2020b}]%
        {pmlr-v119-chen20j}
\bibfield{author}{\bibinfo{person}{Ting Chen}, \bibinfo{person}{Simon
  Kornblith}, \bibinfo{person}{Mohammad Norouzi}, {and}
  \bibinfo{person}{Geoffrey Hinton}.} \bibinfo{year}{2020}\natexlab{b}.
\newblock \showarticletitle{A Simple Framework for Contrastive Learning of
  Visual Representations}. In \bibinfo{booktitle}{\emph{Proceedings of the 37th
  International Conference on Machine Learning}}
  \emph{(\bibinfo{series}{Proceedings of Machine Learning Research},
  Vol.~\bibinfo{volume}{119})}, \bibfield{editor}{\bibinfo{person}{Hal~Daumé
  III} {and} \bibinfo{person}{Aarti Singh}} (Eds.). \bibinfo{publisher}{PMLR},
  \bibinfo{pages}{1597--1607}.
\newblock
\urldef\tempurl%
\url{http://proceedings.mlr.press/v119/chen20j.html}
\showURL{%
\tempurl}


\bibitem[\protect\citeauthoryear{Chen, Fan, Girshick, and He}{Chen
  et~al\mbox{.}}{2020a}]%
        {chen2020improved}
\bibfield{author}{\bibinfo{person}{Xinlei Chen}, \bibinfo{person}{Haoqi Fan},
  \bibinfo{person}{Ross Girshick}, {and} \bibinfo{person}{Kaiming He}.}
  \bibinfo{year}{2020}\natexlab{a}.
\newblock \showarticletitle{Improved baselines with momentum contrastive
  learning}.
\newblock \bibinfo{journal}{\emph{arXiv preprint arXiv:2003.04297}}
  (\bibinfo{year}{2020}).
\newblock


\bibitem[\protect\citeauthoryear{Chum, Philbin, Sivic, Isard, and
  Zisserman}{Chum et~al\mbox{.}}{2007}]%
        {chum2007total}
\bibfield{author}{\bibinfo{person}{Ondrej Chum}, \bibinfo{person}{James
  Philbin}, \bibinfo{person}{Josef Sivic}, \bibinfo{person}{Michael Isard},
  {and} \bibinfo{person}{Andrew Zisserman}.} \bibinfo{year}{2007}\natexlab{}.
\newblock \showarticletitle{Total recall: Automatic query expansion with a
  generative feature model for object retrieval}. In
  \bibinfo{booktitle}{\emph{2007 IEEE 11th International Conference on Computer
  Vision}}. IEEE, \bibinfo{pages}{1--8}.
\newblock


\bibitem[\protect\citeauthoryear{Donahue, Jia, Vinyals, Hoffman, Zhang, Tzeng,
  and Darrell}{Donahue et~al\mbox{.}}{2014}]%
        {donahue2014decaf}
\bibfield{author}{\bibinfo{person}{Jeff Donahue}, \bibinfo{person}{Yangqing
  Jia}, \bibinfo{person}{Oriol Vinyals}, \bibinfo{person}{Judy Hoffman},
  \bibinfo{person}{Ning Zhang}, \bibinfo{person}{Eric Tzeng}, {and}
  \bibinfo{person}{Trevor Darrell}.} \bibinfo{year}{2014}\natexlab{}.
\newblock \showarticletitle{Decaf: A deep convolutional activation feature for
  generic visual recognition}. In \bibinfo{booktitle}{\emph{International
  conference on machine learning}}. PMLR, \bibinfo{pages}{647--655}.
\newblock


\bibitem[\protect\citeauthoryear{Gong, Wang, Guo, and Lazebnik}{Gong
  et~al\mbox{.}}{2014}]%
        {gong2014multi}
\bibfield{author}{\bibinfo{person}{Yunchao Gong}, \bibinfo{person}{Liwei Wang},
  \bibinfo{person}{Ruiqi Guo}, {and} \bibinfo{person}{Svetlana Lazebnik}.}
  \bibinfo{year}{2014}\natexlab{}.
\newblock \showarticletitle{Multi-scale orderless pooling of deep convolutional
  activation features}. In \bibinfo{booktitle}{\emph{European conference on
  computer vision}}. Springer, \bibinfo{pages}{392--407}.
\newblock


\bibitem[\protect\citeauthoryear{Gordo, Almaz{\'a}n, Revaud, and Larlus}{Gordo
  et~al\mbox{.}}{2016}]%
        {gordo2016deep}
\bibfield{author}{\bibinfo{person}{Albert Gordo}, \bibinfo{person}{Jon
  Almaz{\'a}n}, \bibinfo{person}{Jerome Revaud}, {and} \bibinfo{person}{Diane
  Larlus}.} \bibinfo{year}{2016}\natexlab{}.
\newblock \showarticletitle{Deep image retrieval: Learning global
  representations for image search}. In \bibinfo{booktitle}{\emph{European
  conference on computer vision}}. Springer, \bibinfo{pages}{241--257}.
\newblock


\bibitem[\protect\citeauthoryear{Gordo, Almazan, Revaud, and Larlus}{Gordo
  et~al\mbox{.}}{2017}]%
        {gordo2017end}
\bibfield{author}{\bibinfo{person}{Albert Gordo}, \bibinfo{person}{Jon
  Almazan}, \bibinfo{person}{Jerome Revaud}, {and} \bibinfo{person}{Diane
  Larlus}.} \bibinfo{year}{2017}\natexlab{}.
\newblock \showarticletitle{End-to-end learning of deep visual representations
  for image retrieval}.
\newblock \bibinfo{journal}{\emph{International Journal of Computer Vision}}
  \bibinfo{volume}{124}, \bibinfo{number}{2} (\bibinfo{year}{2017}),
  \bibinfo{pages}{237--254}.
\newblock


\bibitem[\protect\citeauthoryear{Gordoa, Rodriguez-Serrano, Perronnin, and
  Valveny}{Gordoa et~al\mbox{.}}{2012}]%
        {gordoa2012leveraging}
\bibfield{author}{\bibinfo{person}{Albert Gordoa}, \bibinfo{person}{Jose~A
  Rodriguez-Serrano}, \bibinfo{person}{Florent Perronnin}, {and}
  \bibinfo{person}{Ernest Valveny}.} \bibinfo{year}{2012}\natexlab{}.
\newblock \showarticletitle{Leveraging category-level labels for instance-level
  image retrieval}. In \bibinfo{booktitle}{\emph{2012 IEEE Conference on
  Computer Vision and Pattern Recognition}}. IEEE, \bibinfo{pages}{3045--3052}.
\newblock


\bibitem[\protect\citeauthoryear{Gutmann and Hyv{\"a}rinen}{Gutmann and
  Hyv{\"a}rinen}{2010}]%
        {gutmann2010noise}
\bibfield{author}{\bibinfo{person}{Michael Gutmann} {and} \bibinfo{person}{Aapo
  Hyv{\"a}rinen}.} \bibinfo{year}{2010}\natexlab{}.
\newblock \showarticletitle{Noise-contrastive estimation: A new estimation
  principle for unnormalized statistical models}. In
  \bibinfo{booktitle}{\emph{Proceedings of the Thirteenth International
  Conference on Artificial Intelligence and Statistics}}. JMLR Workshop and
  Conference Proceedings, \bibinfo{pages}{297--304}.
\newblock


\bibitem[\protect\citeauthoryear{He, Fan, Wu, Xie, and Girshick}{He
  et~al\mbox{.}}{2020}]%
        {he2020momentum}
\bibfield{author}{\bibinfo{person}{Kaiming He}, \bibinfo{person}{Haoqi Fan},
  \bibinfo{person}{Yuxin Wu}, \bibinfo{person}{Saining Xie}, {and}
  \bibinfo{person}{Ross Girshick}.} \bibinfo{year}{2020}\natexlab{}.
\newblock \showarticletitle{Momentum contrast for unsupervised visual
  representation learning}. In \bibinfo{booktitle}{\emph{Proceedings of the
  IEEE/CVF Conference on Computer Vision and Pattern Recognition}}.
  \bibinfo{pages}{9729--9738}.
\newblock


\bibitem[\protect\citeauthoryear{He, Zhang, Ren, and Sun}{He
  et~al\mbox{.}}{2016}]%
        {he2016deep}
\bibfield{author}{\bibinfo{person}{Kaiming He}, \bibinfo{person}{Xiangyu
  Zhang}, \bibinfo{person}{Shaoqing Ren}, {and} \bibinfo{person}{Jian Sun}.}
  \bibinfo{year}{2016}\natexlab{}.
\newblock \showarticletitle{Deep residual learning for image recognition}. In
  \bibinfo{booktitle}{\emph{Proceedings of the IEEE conference on computer
  vision and pattern recognition}}. \bibinfo{pages}{770--778}.
\newblock


\bibitem[\protect\citeauthoryear{Hjelm, Fedorov, Lavoie-Marchildon, Grewal,
  Bachman, Trischler, and Bengio}{Hjelm et~al\mbox{.}}{2018}]%
        {hjelm2018learning}
\bibfield{author}{\bibinfo{person}{R~Devon Hjelm}, \bibinfo{person}{Alex
  Fedorov}, \bibinfo{person}{Samuel Lavoie-Marchildon}, \bibinfo{person}{Karan
  Grewal}, \bibinfo{person}{Phil Bachman}, \bibinfo{person}{Adam Trischler},
  {and} \bibinfo{person}{Yoshua Bengio}.} \bibinfo{year}{2018}\natexlab{}.
\newblock \showarticletitle{Learning deep representations by mutual information
  estimation and maximization}.
\newblock \bibinfo{journal}{\emph{arXiv preprint arXiv:1808.06670}}
  (\bibinfo{year}{2018}).
\newblock


\bibitem[\protect\citeauthoryear{Iscen, Tolias, Avrithis, Furon, and
  Chum}{Iscen et~al\mbox{.}}{2017}]%
        {iscen2017efficient}
\bibfield{author}{\bibinfo{person}{Ahmet Iscen}, \bibinfo{person}{Giorgos
  Tolias}, \bibinfo{person}{Yannis Avrithis}, \bibinfo{person}{Teddy Furon},
  {and} \bibinfo{person}{Ondrej Chum}.} \bibinfo{year}{2017}\natexlab{}.
\newblock \showarticletitle{Efficient diffusion on region manifolds: Recovering
  small objects with compact cnn representations}. In
  \bibinfo{booktitle}{\emph{Proceedings of the IEEE Conference on Computer
  Vision and Pattern Recognition}}. \bibinfo{pages}{2077--2086}.
\newblock


\bibitem[\protect\citeauthoryear{J{\'e}gou, Douze, Schmid, and
  P{\'e}rez}{J{\'e}gou et~al\mbox{.}}{2010}]%
        {jegou2010aggregating}
\bibfield{author}{\bibinfo{person}{Herv{\'e} J{\'e}gou},
  \bibinfo{person}{Matthijs Douze}, \bibinfo{person}{Cordelia Schmid}, {and}
  \bibinfo{person}{Patrick P{\'e}rez}.} \bibinfo{year}{2010}\natexlab{}.
\newblock \showarticletitle{Aggregating local descriptors into a compact image
  representation}. In \bibinfo{booktitle}{\emph{2010 IEEE computer society
  conference on computer vision and pattern recognition}}. IEEE,
  \bibinfo{pages}{3304--3311}.
\newblock


\bibitem[\protect\citeauthoryear{Krizhevsky, Sutskever, and Hinton}{Krizhevsky
  et~al\mbox{.}}{2012}]%
        {krizhevsky2012imagenet}
\bibfield{author}{\bibinfo{person}{Alex Krizhevsky}, \bibinfo{person}{Ilya
  Sutskever}, {and} \bibinfo{person}{Geoffrey~E Hinton}.}
  \bibinfo{year}{2012}\natexlab{}.
\newblock \showarticletitle{Imagenet classification with deep convolutional
  neural networks}.
\newblock \bibinfo{journal}{\emph{Advances in neural information processing
  systems}}  \bibinfo{volume}{25} (\bibinfo{year}{2012}),
  \bibinfo{pages}{1097--1105}.
\newblock


\bibitem[\protect\citeauthoryear{Le-Khac, Healy, and Smeaton}{Le-Khac
  et~al\mbox{.}}{2020}]%
        {le2020contrastive}
\bibfield{author}{\bibinfo{person}{Phuc~H Le-Khac}, \bibinfo{person}{Graham
  Healy}, {and} \bibinfo{person}{Alan~F Smeaton}.}
  \bibinfo{year}{2020}\natexlab{}.
\newblock \showarticletitle{Contrastive representation learning: A framework
  and review}.
\newblock \bibinfo{journal}{\emph{IEEE Access}} (\bibinfo{year}{2020}).
\newblock


\bibitem[\protect\citeauthoryear{Lowe}{Lowe}{2004}]%
        {lowe2004distinctive}
\bibfield{author}{\bibinfo{person}{David~G Lowe}.}
  \bibinfo{year}{2004}\natexlab{}.
\newblock \showarticletitle{Distinctive image features from scale-invariant
  keypoints}.
\newblock \bibinfo{journal}{\emph{International journal of computer vision}}
  \bibinfo{volume}{60}, \bibinfo{number}{2} (\bibinfo{year}{2004}),
  \bibinfo{pages}{91--110}.
\newblock


\bibitem[\protect\citeauthoryear{Mohedano, McGuinness, O'Connor, Salvador,
  Marques, and Gir{\'o}-i Nieto}{Mohedano et~al\mbox{.}}{2016}]%
        {mohedano2016bags}
\bibfield{author}{\bibinfo{person}{Eva Mohedano}, \bibinfo{person}{Kevin
  McGuinness}, \bibinfo{person}{Noel~E O'Connor}, \bibinfo{person}{Amaia
  Salvador}, \bibinfo{person}{Ferran Marques}, {and} \bibinfo{person}{Xavier
  Gir{\'o}-i Nieto}.} \bibinfo{year}{2016}\natexlab{}.
\newblock \showarticletitle{Bags of local convolutional features for scalable
  instance search}. In \bibinfo{booktitle}{\emph{Proceedings of the 2016 ACM on
  International Conference on Multimedia Retrieval}}.
  \bibinfo{pages}{327--331}.
\newblock


\bibitem[\protect\citeauthoryear{Nister and Stewenius}{Nister and
  Stewenius}{2006}]%
        {nister2006scalable}
\bibfield{author}{\bibinfo{person}{David Nister} {and} \bibinfo{person}{Henrik
  Stewenius}.} \bibinfo{year}{2006}\natexlab{}.
\newblock \showarticletitle{Scalable recognition with a vocabulary tree}. In
  \bibinfo{booktitle}{\emph{2006 IEEE Computer Society Conference on Computer
  Vision and Pattern Recognition (CVPR'06)}}, Vol.~\bibinfo{volume}{2}. Ieee,
  \bibinfo{pages}{2161--2168}.
\newblock


\bibitem[\protect\citeauthoryear{Oord, Li, and Vinyals}{Oord
  et~al\mbox{.}}{2018}]%
        {oord2018representation}
\bibfield{author}{\bibinfo{person}{Aaron van~den Oord}, \bibinfo{person}{Yazhe
  Li}, {and} \bibinfo{person}{Oriol Vinyals}.} \bibinfo{year}{2018}\natexlab{}.
\newblock \showarticletitle{Representation learning with contrastive predictive
  coding}.
\newblock \bibinfo{journal}{\emph{arXiv preprint arXiv:1807.03748}}
  (\bibinfo{year}{2018}).
\newblock


\bibitem[\protect\citeauthoryear{Perronnin and Dance}{Perronnin and
  Dance}{2007}]%
        {perronnin2007fisher}
\bibfield{author}{\bibinfo{person}{Florent Perronnin} {and}
  \bibinfo{person}{Christopher Dance}.} \bibinfo{year}{2007}\natexlab{}.
\newblock \showarticletitle{Fisher kernels on visual vocabularies for image
  categorization}. In \bibinfo{booktitle}{\emph{2007 IEEE conference on
  computer vision and pattern recognition}}. IEEE, \bibinfo{pages}{1--8}.
\newblock


\bibitem[\protect\citeauthoryear{Perronnin and Larlus}{Perronnin and
  Larlus}{2015}]%
        {perronnin2015fisher}
\bibfield{author}{\bibinfo{person}{Florent Perronnin} {and}
  \bibinfo{person}{Diane Larlus}.} \bibinfo{year}{2015}\natexlab{}.
\newblock \showarticletitle{Fisher vectors meet neural networks: A hybrid
  classification architecture}. In \bibinfo{booktitle}{\emph{Proceedings of the
  IEEE conference on computer vision and pattern recognition}}.
  \bibinfo{pages}{3743--3752}.
\newblock


\bibitem[\protect\citeauthoryear{Perronnin, Liu, S{\'a}nchez, and
  Poirier}{Perronnin et~al\mbox{.}}{2010}]%
        {perronnin2010large}
\bibfield{author}{\bibinfo{person}{Florent Perronnin}, \bibinfo{person}{Yan
  Liu}, \bibinfo{person}{Jorge S{\'a}nchez}, {and} \bibinfo{person}{Herv{\'e}
  Poirier}.} \bibinfo{year}{2010}\natexlab{}.
\newblock \showarticletitle{Large-scale image retrieval with compressed fisher
  vectors}. In \bibinfo{booktitle}{\emph{2010 IEEE Computer Society Conference
  on Computer Vision and Pattern Recognition}}. IEEE,
  \bibinfo{pages}{3384--3391}.
\newblock


\bibitem[\protect\citeauthoryear{Philbin, Chum, Isard, Sivic, and
  Zisserman}{Philbin et~al\mbox{.}}{2007}]%
        {philbin2007object}
\bibfield{author}{\bibinfo{person}{James Philbin}, \bibinfo{person}{Ondrej
  Chum}, \bibinfo{person}{Michael Isard}, \bibinfo{person}{Josef Sivic}, {and}
  \bibinfo{person}{Andrew Zisserman}.} \bibinfo{year}{2007}\natexlab{}.
\newblock \showarticletitle{Object retrieval with large vocabularies and fast
  spatial matching}. In \bibinfo{booktitle}{\emph{2007 IEEE conference on
  computer vision and pattern recognition}}. IEEE, \bibinfo{pages}{1--8}.
\newblock


\bibitem[\protect\citeauthoryear{Philbin, Chum, Isard, Sivic, and
  Zisserman}{Philbin et~al\mbox{.}}{2008}]%
        {philbin2008lost}
\bibfield{author}{\bibinfo{person}{James Philbin}, \bibinfo{person}{Ondrej
  Chum}, \bibinfo{person}{Michael Isard}, \bibinfo{person}{Josef Sivic}, {and}
  \bibinfo{person}{Andrew Zisserman}.} \bibinfo{year}{2008}\natexlab{}.
\newblock \showarticletitle{Lost in quantization: Improving particular object
  retrieval in large scale image databases}. In \bibinfo{booktitle}{\emph{2008
  IEEE conference on computer vision and pattern recognition}}. IEEE,
  \bibinfo{pages}{1--8}.
\newblock


\bibitem[\protect\citeauthoryear{Radenovi{\'c}, Iscen, Tolias, Avrithis, and
  Chum}{Radenovi{\'c} et~al\mbox{.}}{2018}]%
        {radenovic2018revisiting}
\bibfield{author}{\bibinfo{person}{Filip Radenovi{\'c}}, \bibinfo{person}{Ahmet
  Iscen}, \bibinfo{person}{Giorgos Tolias}, \bibinfo{person}{Yannis Avrithis},
  {and} \bibinfo{person}{Ond{\v{r}}ej Chum}.} \bibinfo{year}{2018}\natexlab{}.
\newblock \showarticletitle{Revisiting oxford and paris: Large-scale image
  retrieval benchmarking}. In \bibinfo{booktitle}{\emph{Proceedings of the IEEE
  Conference on Computer Vision and Pattern Recognition}}.
  \bibinfo{pages}{5706--5715}.
\newblock


\bibitem[\protect\citeauthoryear{Radenovi{\'c}, J{\'e}gou, and
  Chum}{Radenovi{\'c} et~al\mbox{.}}{2015}]%
        {radenovic2015multiple}
\bibfield{author}{\bibinfo{person}{Filip Radenovi{\'c}},
  \bibinfo{person}{Herv{\'e} J{\'e}gou}, {and} \bibinfo{person}{Ondrej Chum}.}
  \bibinfo{year}{2015}\natexlab{}.
\newblock \showarticletitle{Multiple measurements and joint dimensionality
  reduction for large scale image search with short vectors}. In
  \bibinfo{booktitle}{\emph{Proceedings of the 5th ACM on International
  Conference on Multimedia Retrieval}}. \bibinfo{pages}{587--590}.
\newblock


\bibitem[\protect\citeauthoryear{Radenovi{\'c}, Tolias, and Chum}{Radenovi{\'c}
  et~al\mbox{.}}{2016}]%
        {radenovic2016cnn}
\bibfield{author}{\bibinfo{person}{Filip Radenovi{\'c}},
  \bibinfo{person}{Giorgos Tolias}, {and} \bibinfo{person}{Ond{\v{r}}ej Chum}.}
  \bibinfo{year}{2016}\natexlab{}.
\newblock \showarticletitle{CNN image retrieval learns from BoW: Unsupervised
  fine-tuning with hard examples}. In \bibinfo{booktitle}{\emph{European
  conference on computer vision}}. Springer, \bibinfo{pages}{3--20}.
\newblock


\bibitem[\protect\citeauthoryear{Razavian, Azizpour, Sullivan, and
  Carlsson}{Razavian et~al\mbox{.}}{2014}]%
        {razavian2014cnn}
\bibfield{author}{\bibinfo{person}{Ali~Sharif Razavian},
  \bibinfo{person}{Hossein Azizpour}, \bibinfo{person}{Josephine Sullivan},
  {and} \bibinfo{person}{Stefan Carlsson}.} \bibinfo{year}{2014}\natexlab{}.
\newblock \showarticletitle{CNN features off-the-shelf: an astounding baseline
  for recognition. 2014}.
\newblock \bibinfo{journal}{\emph{arXiv preprint arXiv:1403.6382}}
  (\bibinfo{year}{2014}).
\newblock


\bibitem[\protect\citeauthoryear{Ren, He, Girshick, and Sun}{Ren
  et~al\mbox{.}}{2016}]%
        {ren2016faster}
\bibfield{author}{\bibinfo{person}{Shaoqing Ren}, \bibinfo{person}{Kaiming He},
  \bibinfo{person}{Ross Girshick}, {and} \bibinfo{person}{Jian Sun}.}
  \bibinfo{year}{2016}\natexlab{}.
\newblock \showarticletitle{Faster R-CNN: towards real-time object detection
  with region proposal networks}.
\newblock \bibinfo{journal}{\emph{IEEE transactions on pattern analysis and
  machine intelligence}} \bibinfo{volume}{39}, \bibinfo{number}{6}
  (\bibinfo{year}{2016}), \bibinfo{pages}{1137--1149}.
\newblock


\bibitem[\protect\citeauthoryear{Revaud, Almaz{\'{a}}n, de~Rezende, and
  de~Souza}{Revaud et~al\mbox{.}}{2019}]%
        {DBLP:journals/corr/abs-1906-07589}
\bibfield{author}{\bibinfo{person}{J{\'{e}}r{\^{o}}me Revaud},
  \bibinfo{person}{Jon Almaz{\'{a}}n}, \bibinfo{person}{Rafael~Sampaio de
  Rezende}, {and} \bibinfo{person}{C{\'{e}}sar~Roberto de Souza}.}
  \bibinfo{year}{2019}\natexlab{}.
\newblock \showarticletitle{Learning with Average Precision: Training Image
  Retrieval with a Listwise Loss}.
\newblock \bibinfo{journal}{\emph{CoRR}}  \bibinfo{volume}{abs/1906.07589}
  (\bibinfo{year}{2019}).
\newblock
\showeprint[arxiv]{1906.07589}
\urldef\tempurl%
\url{http://arxiv.org/abs/1906.07589}
\showURL{%
\tempurl}


\bibitem[\protect\citeauthoryear{Russakovsky, Deng, Su, Krause, Satheesh, Ma,
  Huang, Karpathy, Khosla, Bernstein, et~al\mbox{.}}{Russakovsky
  et~al\mbox{.}}{2015}]%
        {russakovsky2015imagenet}
\bibfield{author}{\bibinfo{person}{Olga Russakovsky}, \bibinfo{person}{Jia
  Deng}, \bibinfo{person}{Hao Su}, \bibinfo{person}{Jonathan Krause},
  \bibinfo{person}{Sanjeev Satheesh}, \bibinfo{person}{Sean Ma},
  \bibinfo{person}{Zhiheng Huang}, \bibinfo{person}{Andrej Karpathy},
  \bibinfo{person}{Aditya Khosla}, \bibinfo{person}{Michael Bernstein},
  {et~al\mbox{.}}} \bibinfo{year}{2015}\natexlab{}.
\newblock \showarticletitle{Imagenet large scale visual recognition challenge}.
\newblock \bibinfo{journal}{\emph{International journal of computer vision}}
  \bibinfo{volume}{115}, \bibinfo{number}{3} (\bibinfo{year}{2015}),
  \bibinfo{pages}{211--252}.
\newblock


\bibitem[\protect\citeauthoryear{Simonyan and Zisserman}{Simonyan and
  Zisserman}{2014}]%
        {simonyan2014very}
\bibfield{author}{\bibinfo{person}{Karen Simonyan} {and}
  \bibinfo{person}{Andrew Zisserman}.} \bibinfo{year}{2014}\natexlab{}.
\newblock \showarticletitle{Very deep convolutional networks for large-scale
  image recognition}.
\newblock \bibinfo{journal}{\emph{arXiv preprint arXiv:1409.1556}}
  (\bibinfo{year}{2014}).
\newblock


\bibitem[\protect\citeauthoryear{Sivic and Zisserman}{Sivic and
  Zisserman}{2003}]%
        {sivic2003video}
\bibfield{author}{\bibinfo{person}{Josef Sivic} {and} \bibinfo{person}{Andrew
  Zisserman}.} \bibinfo{year}{2003}\natexlab{}.
\newblock \showarticletitle{Video Google: A text retrieval approach to object
  matching in videos}. In \bibinfo{booktitle}{\emph{Computer Vision, IEEE
  International Conference on}}, Vol.~\bibinfo{volume}{3}. IEEE Computer
  Society, \bibinfo{pages}{1470--1470}.
\newblock


\bibitem[\protect\citeauthoryear{Tolias, Sicre, and J{\'e}gou}{Tolias
  et~al\mbox{.}}{2015}]%
        {tolias2015particular}
\bibfield{author}{\bibinfo{person}{Giorgos Tolias}, \bibinfo{person}{Ronan
  Sicre}, {and} \bibinfo{person}{Herv{\'e} J{\'e}gou}.}
  \bibinfo{year}{2015}\natexlab{}.
\newblock \showarticletitle{Particular object retrieval with integral
  max-pooling of CNN activations}.
\newblock \bibinfo{journal}{\emph{arXiv preprint arXiv:1511.05879}}
  (\bibinfo{year}{2015}).
\newblock


\bibitem[\protect\citeauthoryear{Wang and Jiang}{Wang and Jiang}{2015}]%
        {wang2015instre}
\bibfield{author}{\bibinfo{person}{Shuang Wang} {and} \bibinfo{person}{Shuqiang
  Jiang}.} \bibinfo{year}{2015}\natexlab{}.
\newblock \showarticletitle{Instre: a new benchmark for instance-level object
  retrieval and recognition}.
\newblock \bibinfo{journal}{\emph{ACM Transactions on Multimedia Computing,
  Communications, and Applications (TOMM)}} \bibinfo{volume}{11},
  \bibinfo{number}{3} (\bibinfo{year}{2015}), \bibinfo{pages}{1--21}.
\newblock


\bibitem[\protect\citeauthoryear{Wu, Xiong, Yu, and Lin}{Wu
  et~al\mbox{.}}{2018}]%
        {wu2018unsupervised}
\bibfield{author}{\bibinfo{person}{Zhirong Wu}, \bibinfo{person}{Yuanjun
  Xiong}, \bibinfo{person}{Stella~X Yu}, {and} \bibinfo{person}{Dahua Lin}.}
  \bibinfo{year}{2018}\natexlab{}.
\newblock \showarticletitle{Unsupervised feature learning via non-parametric
  instance discrimination}. In \bibinfo{booktitle}{\emph{Proceedings of the
  IEEE Conference on Computer Vision and Pattern Recognition}}.
  \bibinfo{pages}{3733--3742}.
\newblock


\bibitem[\protect\citeauthoryear{Yang, Hinami, Matsui, Ly, and Satoh}{Yang
  et~al\mbox{.}}{2019}]%
        {yang2019efficient}
\bibfield{author}{\bibinfo{person}{Fan Yang}, \bibinfo{person}{Ryota Hinami},
  \bibinfo{person}{Yusuke Matsui}, \bibinfo{person}{Steven Ly}, {and}
  \bibinfo{person}{Shin’ichi Satoh}.} \bibinfo{year}{2019}\natexlab{}.
\newblock \showarticletitle{Efficient image retrieval via decoupling diffusion
  into online and offline processing}. In \bibinfo{booktitle}{\emph{Proceedings
  of the AAAI Conference on Artificial Intelligence}},
  Vol.~\bibinfo{volume}{33}. \bibinfo{pages}{9087--9094}.
\newblock


\end{thebibliography}
\end{document}